\documentclass{article}

\PassOptionsToPackage{numbers, sort&compress}{natbib}
\usepackage[preprint]{neurips_2023}

\usepackage[utf8]{inputenc} 
\usepackage[T1]{fontenc}    
\usepackage{url}            
\usepackage{booktabs}       
\usepackage{amsfonts}       
\usepackage{nicefrac}       
\usepackage{microtype}      
\usepackage{enumitem}
\usepackage{wrapfig} 
\usepackage{graphicx}
\usepackage{subfigure}
\usepackage{amsmath}
\usepackage{amssymb}
\usepackage{booktabs}
\usepackage{bbm}
\usepackage[dvipsnames,svgnames,x11names,table]{xcolor} 
\usepackage{pifont}
\usepackage{bbding}

\usepackage{wrapfig}
\usepackage{multirow}
\usepackage{transparent}
\usepackage{tikz}

\definecolor{pretty-blue}{RGB}{0, 113, 188}
\definecolor{brown}{RGB}{201, 104, 71}

\usepackage[colorlinks=True,
            linkcolor=blue,
            anchorcolor=blue,  
            pagebackref,
            urlcolor=blue,
            citecolor=teal,
            ]{hyperref}

\title{\ \ \ \ Small Language Model Meets with \\ \ \ Reinforced Vision Vocabulary}

\author{
Haoran Wei$^{1,}$\thanks{Equal contribution} \quad
Lingyu Kong$^{2,*}$ \quad Jinyue Chen$^{2}$ \quad Liang Zhao$^{1}$ \\  \bf{Zheng Ge}$^{1}$\thanks{Project leader} \quad \bf{En Yu}$^{3}$ \quad Jianjian Sun$^{1}$ \quad Chunrui Han$^{1}$ \quad Xiangyu Zhang$^{1}$\\
$^{1}$MEGVII Technology \ \  $^{2}$University of Chinese Academy of Sciences  \\ $^{3}$Huazhong University of Science and Technology\\
{\url{https://varytoy.github.io/}}
}

\begin{document}

\maketitle





\begin{abstract}
Playing Large Vision Language Models (LVLMs) in 2023 is trendy among the AI community. However, the relatively large number of parameters (more than 7B) of popular LVLMs makes it difficult to train and deploy on consumer GPUs, discouraging many researchers with limited resources. Imagine how cool it would be to experience all the features of current LVLMs on an old GTX1080ti (our only game card).  Accordingly, we present Vary-toy in this report, a small-size Vary along with Qwen-1.8B as the base ``large'' language model. In Vary-toy, we introduce an improved vision vocabulary, allowing the model to not only possess all features of Vary but also gather more generality. Specifically, we replace negative samples of natural images with positive sample data driven by object detection in the procedure of generating vision vocabulary, more sufficiently utilizing the capacity of the vocabulary network and enabling it to efficiently encode visual information corresponding to natural objects. For experiments, Vary-toy can achieve 65.6\% ANLS on DocVQA, 59.1\% accuracy on ChartQA, 88.1\% accuracy on RefCOCO, and 29\% on MMVet.   The code will be publicly available on the homepage.

\end{abstract}

\section{Introduction}
\label{intro}
Large Vision Language Model (LVLM)  is one of the hottest research topics~\cite{BLIP2,Flamingo,llava,minigpt4,InstructGPT,wei2023vary} in the field of artificial intelligence among the last year. The exciting part is that one LVLM can achieve satisfactory performance in many downstream tasks~\cite{COCO,TextVQA,coco_text,OCRVQA,STVQA,DocVQA} guided by different prompts. However, there is still significant room for improvement in LVLM's overall image perception capacity. Intuitively, an advanced perceptual ability for visual concepts is essential to enhance the further development and implementation of a model. We deem that there are two main challenges to achieve that: 1) the shortcomings of the current vision vocabulary network~\cite{radford2021learning,wei2023vary} in extracting rich visual information; 2) the huge model iteration cost in the optimization of a large number of parameters.

\begin{figure}[t]
\centering
\includegraphics[width=1.0\textwidth]{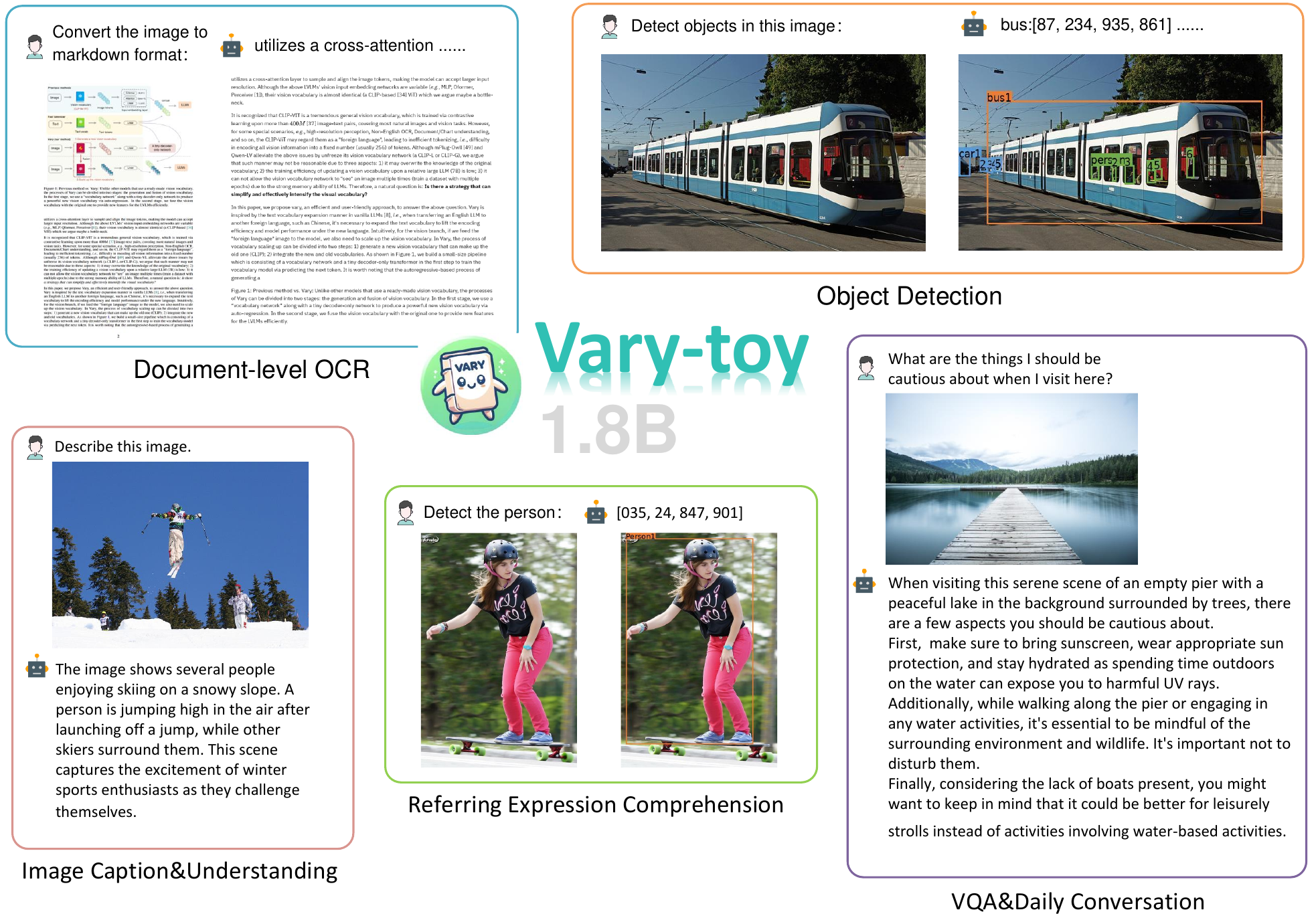}
\caption{Features of Vary-toy. Based on a 1.8B language model, Vary-toy can achieve all features of vanilla Vary-base, including document OCR, image caption, VQA, general conversation, and so on. Besides, we introduce the natural object perception (location) ability for Vary-toy. Most importantly, with just only a single GTX1080ti GPU, you can experience all of the above.}
\label{fig1}
\end{figure}

As aforementioned, current LVLMs demonstrate amazing ability in many tasks, especially the Computer Vision (CV) and Natural Language Processing (NLP) intersected ones (\textit{e.g.,} image capion~\cite{COCO}, VQA~\cite{TextVQA}, memes understanding, scene OCR~\cite{OCRVQA}, \textit{etc}), based on the almost perfect vision vocabulary network — CLIP~\cite{radford2021learning}.  The structures of popular LVLMs can be divided into two main streams:  1) image tokens as prefixes like MetaLM~\cite{hao2022language}; 2) cross-attention for feature fusion like Flamingo~\cite{Flamingo}. Regardless of which structure is used, the upper limit of the model may be hindered by the visual signals encoding efficiency of its vision vocabulary network. To break through the potential bottleneck, Vary~\cite{wei2023vary} introduces a simple and effective manner to scale up the vision vocabulary for an LVLM. The scaling law is to first train a new visual vocabulary network using a small auto-regressive model (OPT-125M~\cite{OPT}), and then merge the old and new vocabularies to form the final LVLM (Vary-base~\cite{wei2023vary}). However, Vary suffers two drawbacks to being a user-friendly baseline: 1) The waste of network capacity in the new vision vocabulary (which in vanilla Vary is only used to compress text information in PDF images).  2) The Vary-base with 7B LLM takes high iteration costs (requiring multiple A100 machines to train).

In this report, we present a small-size Vary, \textit{i.e.,} Vary-toy, to alleviate the aforementioned issues. Overall, Vary-toy enjoys the same pipeline as vanilla Vary, including a vision vocabulary generating and scaling up processes. Considering the original Vary masks natural images as negative samples during the creation of a new visual vocabulary. We believe this procedure, to some extent, wastes network capacity, leaving room for optimization. Instead, we regard the natural image as the object detection task~\cite{ren2015faster,redmon2016you,lin2017focal,law2018cornernet,zhou2019bottom,ijcai2022p203,carion2020end}. Thus in processing the vision vocabulary, we incorporate both dense textual data (PDF) and natural object location data into the vocabulary network of Vary-toy, making it more universal.  After completing the new and reinforced vocabulary, we merge it with the genuine (224$\times$224) CLIP and then integrate them into a 1.8B language model~\cite{qwen}.

In experiments, we report metrics on several challenging benchmarks, \textit{i.e.,} DocVQA~\cite{DocVQA}, ChartQA~\cite{masry2022chartqa}, MMvet~\cite{yu2023mm}, and RefCOCO~\cite{kazemzadeh2014referitgame}. Specifically, Vary-toy can achieve 65.6\% ANLS on DocVQA, 59.1\% accuracy on ChartQA, 29\% accuracy on MMvet, and 88.1\% accuracy on RefCOCO val. More specifically, it can gather on par performance compared to Qwen-VL-7B~\cite{Qwen-VL} on DocVQA and RefCOCO as well as a better accuracy than LLaVA-7B~\cite{llava} on the general benchmark MMVet. 

In conclusion, Vary-toy is a toy because it is at least three times smaller compared to popular LVLMs (>7B). Vary-toy is not a toy due to it demonstrates excellent potential in challenging tasks. We believe that Vary-toy still enjoys many improvement rooms and we hope that our small-size LVLM can encourage more attention in corresponding research and become a practical baseline, especially for those researchers with limited resources.

\begin{figure}[t]
\centering
\includegraphics[width=1.0\textwidth]{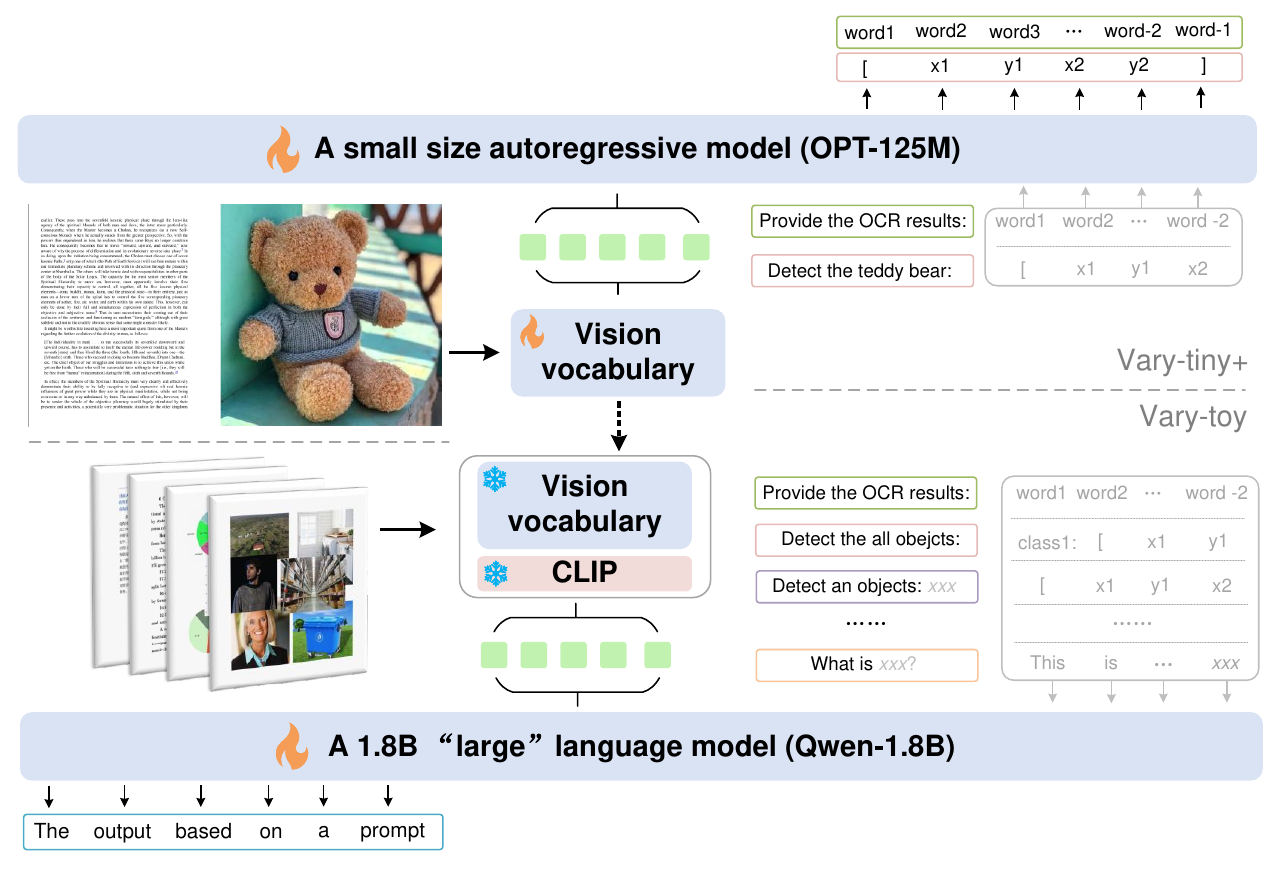}
\caption{Architecture of the Vary-toy. We utilize the Vary-tiny+ pipeline to generate the new vision vocabulary of Vary-toy. Such vision vocabulary can efficiently encode dense text and natural object location information into tokens. Based on the improved vocabulary, Vary-toy not only possesses all the previous features (document OCR) but also handles object detection tasks well.}
\label{fig2}
\end{figure}

\section{Related Works}
Over the past years, Large Language Models (LLMs), such as the GPT family~\cite{GPT-2,GPT3,InstructGPT}, LLaMA family~\cite{llama,alpaca,vicuna}, OPT~\cite{OPT}, and the GLM family~\cite{GLM} gain significantly advanced performance in NLP tasks. With the help of LLMs’ language reasoning abilities, Vision Language Models (VLMs) like Flamingo~\cite{Flamingo}, BLIP2~
 \cite{BLIP2}, LLaVA~\cite{llava,liu2023improvedllava}, Vary~\cite{wei2023vary}, \textit{etc}~\cite{minigpt4,Qwen-VL,dong2023dreamllm,zhao2023chatspot,yu2023merlin} have achieved impressive results in various computer vision tasks such as image caption~\cite{COCO}, VQA~\cite{DocVQA,STVQA,OCRVQA}, image generation~\cite{dong2023dreamllm}, visual grounding~\cite{minigpt4,Qwen-VL,yu2023merlin}, document OCR~\cite{wei2023vary} and so on. These models not only can follow human instructions but also possess remarkable few-shot and even zero-shot learning abilities, thereby driving the AI community toward the development of artificial general intelligence (AGI).

However, most popular open-source VLMs are parameter-heavy, with sizes like 7B (\textit{e.g.,} Qwen-VL~\cite{Qwen-VL} and mPlUG-Owl~\cite{ye2023mplug}) or 13B~\cite{llava}, which to some extend hinder the participation of researchers with limited resources and poses challenges for the implementation of VLMs in resource-constrained environments like home computer. Recently, there has been a growing interest in and development of smaller language models, such as Phi-2 (2.7B)~\cite{phi-2} and Qwen-1.8B~\cite{qwen} for NLP tasks, and Gemini-nano (1.8B/3.25B)~\cite{team2023gemini}, MobileVLM (1.4B/2.7B)~\cite{chu2023mobilevlm} for vision-language tasks.

In this report, Vary-toy will be an open-source small model that possesses features of the most popular LVLMs and demonstrates exceptional potential in fine-grained perception tasks.

\section{Method}
In this section, we will delve into the details of how to devise Vary-toy.  As shown in Figure~\ref{fig2}, there are two main parts in implementing the model:  1) how to generate a more practical vision vocabulary based on the Vary-tiny+ pipeline. 2) how to utilize the new vision vocabulary to make the 1.8B Vary-toy gather new features on the premise of not harming the original model features. 

\subsection{Generating A Reinforced Vision Vocabulary Upon Vary-tiny+}
Vary-tiny~\cite{wei2023vary} is a tiny vision language model to generate a specific PDF-parsing vision vocabulary for Vary. The vision vocabulary network comprises a SAM-base~\cite{kirillov2023segment} main body and paired convolutions to reshape the output, enjoying about 80M parameters. Experiments in Vary prove that using the SAM initializing to gain intensive text perception is effective. However, the vocabulary-generating procedure in vanilla Vary suffers the risk of forgetting SAM's original natural object perception ability. What's more, we also think that writing only the visual knowledge of dense text into an 80M network is wasteful.  Thus we generate a new and more reasonable vision vocabulary upon the Vary-tiny+ pipeline.

\begin{figure}[h]
\centering
\includegraphics[width=1.0\textwidth]{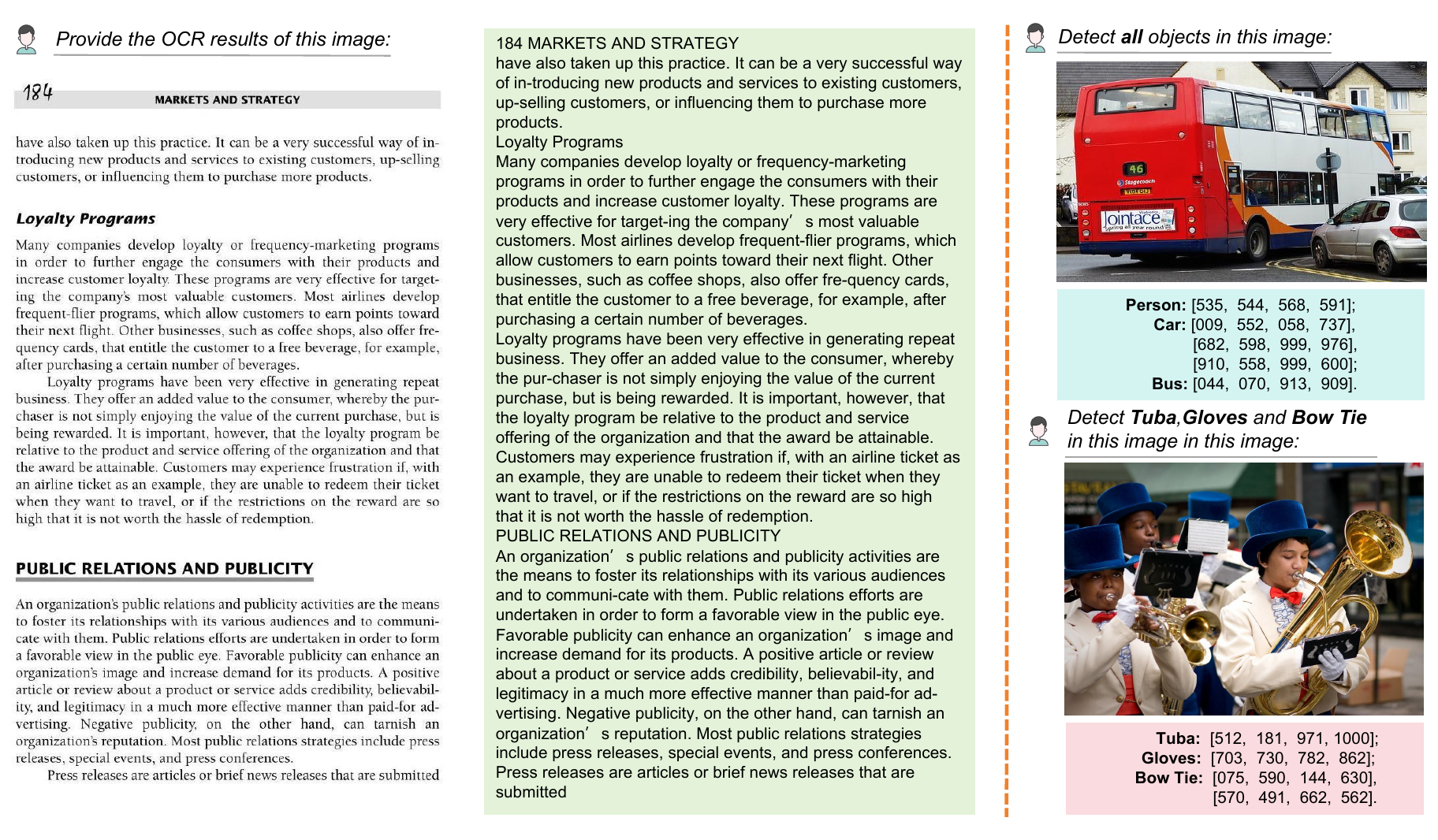}
\caption{Visualization of image-text pairs used by Vary-tiny+. For PDF image-text pair, there is only one prompt, while for the object detection task, we utilize two types of prompts as shown in the right half of the figure because some images may have too many objects that exceed the maximum token length (4096) of the OPT125M after interpolation.}
\label{fig3}
\end{figure}

\subsubsection{Data Engine} 
\textbf{PDF data.}  We prepare about 4M PDF image-text pairs in this stage.  Following Vary, we use the PDF processing packages to extract the texts of each PDF page, which we find many Python packages can realize (\textit{e.g.,} pdfminer, pdfplumber, and fitz).  Each page will be saved as a JPEG image and form an image-text pair with the corresponding text. In this way, we get 2M samples for English and 2M for Chinese. We use the sentence: ``Provide the OCR results of this image.'' as the prompt for both English and Chinese tasks. The PDFs are mainly from arXiv, CC-MAIN-2021-31-PDF-UNTRUNCATED, and e-books. Figure~\ref{fig3} shows a sample of the PDF image-pair.

\textbf{Object detection data.} To fully utilize the capacity of the visual vocabulary network and obtain the natural image perception ability from SAM initialization, we introduce object detection data in the vision vocabulary generating process. We gather the samples from two large open-source datasets, \textit{i.e.,} Object365~\cite{objects365} and OpenImage~\cite{kuznetsova2020open}.  Due to the low efficiency of coordinate (number texts) encoding in OPT's~\cite{OPT} text tokenizer, for images with too many objects, the number of tokens in the ground truth may exceed the maximum token length supported by OPT-125M (although we interpolate it to 4096). Therefore, we re-organize the annotations into two tasks: 1)  \textbf{Object Detection}:  If there are no more than 30 object-boxes in the image, we will allow the Vary-tiny+ detect all objects with the prompt: ``\textit{Detect all objects in this image}''. 2) \textbf{REC}:  If the object-box number is over 30, we will regard this image as a REC task using a prompt template: ``\textit{Detect class1, class2, ..., in this image}''. The selected classes are random so one image can be used multiple times. Through the above manner, we obtain approximately 3M of detection data. Some samples can be seen in Figure~\ref{fig3}.

\subsubsection{Input Format} 
Different from the single input/output form of Vary-tiny, Vary-tiny+ needs various input formats to adapt to corresponding tasks due to it requires different prompts to guide the model output correct results. For simplicity, we use the template of Vicuna v1~\cite{vicuna} to construct all ground truth in a conversation format as USER: <img>"<image>"</img> "\textit{texts input}" ASSITANT: "\textit{texts output}" </s>. We add the "<img>" and "</img>" as special tokens of the text tokenizer of OPT-125M and we find that it can adapt very well to the Vicuna template. For the vision input branch, we don't utilize any augmentations and only resize the image to a fixed resolution, \textit{i.e.,} 1024$\times$1024.

\subsection{Forge the Cost-Effective Vary-Toy}
In this section, we depict the design details of Vary-toy, mainly including the structure of the network and the data construction utilized in the pre-training and SFT stages.  

\subsubsection{Architecture}

As shown in Figure~\ref{fig2}, we follow the Vary pipeline to devise the main body of Vary-toy but there are some minor differences.  When fed an input image with a shape of H$\times$W, the new vision vocabulary branch will directly resize the image to  1024$\times$1024, while the CLIP~\cite{radford2021learning} branch gains a 224$\times$224 image by the center crop. Both the two branches output 256 tokens with channels of 1024. The dimension of the Qwen-1.8B's input channel is also 2048, so the simplest manner is to concatenate the image tokens in two branches directly as the input image tokens of the language model. In terms of code implementation, to maintain consistency with the Vary structure, we still add input embedding layers behind the vision vocabulary networks.

\begin{table}[!h]\normalsize
	\centering	
	\setlength{\tabcolsep}{3.0mm}{	
		
		\begin{tabular}{l|l|c|c}
			\toprule 
			\textbf{Task}&\textbf{Dataset} & \textbf{Sample} & \textbf{A prompt example}\\ 
			\midrule  
			\multirow{2}{*}{Cap.} &Laion-COCO~\cite{schuhmann2021laion} & 4M & Describe the content of this      
                                       image in a sentence. \\
			                         &BLIP558k~\cite{llava} & 558K & Describe the image with one saying. \\
                \midrule
			\multirow{2}{*}{PDF}  &Pure OCR & 1M  & Provide the OCR results of this image.  \\
			                       &Markdown& 500K & Convert the image to markdown format. \\
                \midrule
			\multirow{2}{*}{Det.} &COCO~\cite{COCO}&50K&Detect all objects in this image.    \\
			                       &RefCOCO&train set&Detect an object: the left woman.   \\

                \midrule
                \multirow{3}{*}{NLP}  &ShareGPT&125K& \textit{Original conversation}   \\
			                       &Baize~\cite{xu2023baize}&112K&\textit{Original conversation}   \\
                                      &Alpaca~\cite{alpaca}&52K&\textit{Original conversation}   \\
                \midrule
			\multirow{2}{*}{VQA} &DocVQA~\cite{DocVQA}&train set&\textit{Qestion}.Answer using a single word or phrase. \\
			                       &ChartVQA~\cite{masry2022chartqa}&train set&\textit{Qestion}.Answer using a single-word 
                                      or phrase.   \\
  
			\bottomrule		
	\end{tabular}}	
        \vspace{3mm}
 	\caption{Multi-task training data. We introduce 5 types of data in the pretrain stage, including weakly supervised pair data, PDF image-text pair data, detection data, pure text auto-regressive data, and VQA data. All data annotations are reorganized to a conversation format.}
	\label{table1}
\end{table}

\subsubsection{Data Details}

Intuitively, the sensitivity of the 1.8B model to data quantity and ratio is higher than that of the 7B or above models, so we put more effort into the data processing aspect for Vary-toy.

\textbf{Pre-training \& SFT data.} For Vary-toy, the pretrain stage is actually a multi-task training stage, wherein we prepare a large amount of image-text pairs in various formats. As summarized in Table~\ref{table1}, we mainly focus on a total of 5 types of data in such stage, containing weakly annotated image caption, PDF dense OCR, object detection, pure text conversation, and VQA. Specifically, for natural images, we sample 4M image-text pair in the Laion-COCO~\cite{schuhmann2021laion} dataset, and we also use the BLIP-558K data proposed in LLaVA~\cite{llava}.  For PDF image-text pair, we prepare two types of data following Vary. One is pure dense text OCR, and the other is a task that converts the PDF image to a markdown format. The previous type of data is randomly sampled from the PDF data used in Vary-tiny+ and the last one is obtained via \textit{LaTeX} rendering. Compared to vanilla Vary, we reduce the proportion of PDF data to maintain universal capability. For the detection data, we gather images from the COCO~\cite{COCO} dataset. We sample 50K images with fewer objects included for the pure object detection task and use all train data of RefCOCO for the REC task. We normalize the coordinates of each box and then magnify them to 1000 times. To prevent the language ability of the LLM from deteriorating, we also introduce pure NLP conversation data, including ShareGPT, Baize~\cite{xu2023baize}, and Alpaca~\cite{alpaca}.  For the last downstream VQA tasks, we choose two challenge datasets (DocVQA and ChartQA~\cite{masry2022chartqa}) to monitor the text perception and reasoning performance of Vary-toy for artificial data. There are at least 10 prompts made through GPT3.5~\cite{GPT3} for each task, and Table~\ref{table1} shows one example of them.

In the SFT stage, we only use the LLaVA-80K~\cite{llava} to instruction tuning the model. LLaVA-80K is a dataset with detailed descriptions and prompts of various types of images, produced by GPT4~\cite{llava,GPT4}.

\subsubsection{Data Format}
In Vary-toy, we are pleased to keep the Chinese PDF-parsing feature to some extent because there is very little exploration in this area, which is also one of the reasons that we select Qwen-1.8B~\cite{qwen} as our base language model (due to the relatively comprehensive text vocabulary). The data input to Qwen-1.8B follows the vanilla Vary~\cite{wei2023vary} format. That is: <|im\_start|>user: <img>"<image>"</img> "\textit{human prompts}"<|im\_end|> <|im\_start|>assistant: "\textit{model outputs}" <|im\_end|>. 

\section{Experiments} \label{exp}
\subsection{Evaluation Metrics}
We report the accuracy of Vary-toy on four popular and challenging benchmarks: DocVQA~\cite{DocVQA}, ChartQA~\cite{masry2022chartqa}, RefCOCO~\cite{kazemzadeh2014referitgame}, and MMVet~\cite{yu2023mm}. Wherein, the DocVQA and ChartQA can measure the text perception and reasoning ability of the model in manual images, RefCOCO can be used to test the model's ability to locate natural objects, while MMVet, including 6 measurement areas, can be utilized to monitor the general ability of Vary-toy. We use the evaluation metrics introduced in their original paper for fair comparison. Specifically, we utilize ANLS, relaxed accuracy, accuracy under 0.5 IoU, and GPT4 scoring as the metrics for the above four datasets.

\subsection{Implementation Details}

For Vary-tiny+, we unfreeze all the parameters and train the whole model with a batch size of 512 for 2 epochs. We select the AdamW~\cite{AdamW} optimizer with a cosine annealing scheduler~\cite{loshchilov2016sgdr}. The initial learning rate is set to 5e-5 and the end is 0. It is worth noting that the Vary-tiny is initialized by the weights of Vary-tiny for faster convergence.

For Vary-toy, following vanilla Vary, we freeze all weights of two vision vocabulary networks and only optimize the parameters of the input embedding layers and language model (Qwen-1.8B). In the multi-task training (pre-training) stage, we set the start learning rate to be 5e-5 while it is set to 2e-5 in SFT.  We train the model with a batch size of 512 for only 1 epoch in both two stages.

\begin{table}[!h]
        \centering
	\begin{tabular}{lcccccc}
        \toprule[.9pt]
        \multirow{3}{*}{\textbf{Method}} & \multirow{3}{*}{\textbf{Size}}  & \multicolumn{2}{c}{DocVQA} &\multicolumn{3}{c}{ChartQA} \\
        \cmidrule(rl){3-4}  \cmidrule(rl){5-7}
            & & \textbf{val} & \textbf{test} & \textbf{human} & \textbf{augmented} & \textbf{Average}  \\
            \midrule
		Dessurt~\cite{davis2022end}  &-&  46.5 & 63.2 & -    & -    & - \\
            Donut~\cite{kim2022ocr} & -& - & 67.5 & - & - & 41.8\\
            Pix2Sturct~\cite{lee2023pix2struct} & - &   - &    72.1 & 30.5 & 81.6 & 56.0 \\
            mPLUG-DocOwl~\cite{ye2023mplug}  & 7B  & 62.2 & - & - &- & 57.4 \\
            Qwen-VL-chat~\cite{qwen}   &  7B & 65.1 & - & - &- & 65.7 \\

            \midrule
            Vary-toy & 1.8B &  65.6 & 65.0 & 33.4 & 84.8 & 59.1 \\
        \bottomrule[.9pt]
	\end{tabular}
        \setlength{\abovecaptionskip}{0.2cm}
        \vspace{1mm}
       \caption{Performance comparison to popular methods on DocVQA and ChartQA. Vary-toy can achieve 65.6\% ANLS on DocVQA which is on par with the 7B Qwen-VL-chat and 59.1\% accuracy on ChartQA which is higher than 7B-size mPLUG-DocOwl.}
        \label{tab:2}
        %
\end{table}

\subsection{Manual Image Understanding Ability}
We evaluate the fine-grained text perception and reasoning ability via the DocVQA~\cite{DocVQA} and ChartQA~\cite{masry2022chartqa}. As shown in Table~\ref{tab:2}, along with the only 1.8B language model, Vary-toy can achieve 65.6\% ANLS on DocVQA and 59.1\% accuracy on ChartQA. For DocVQA, the Vary-toy enjoys comparable performance to the 7B-size Qwen-VL-chat, proving the excellent document-level text perception ability of the model and also proving that the new vision vocabulary is available on tokenizing PDF images. For ChartQA, Vary-toy can achieve 59.1\% average accuracy, which is better than the 7B size mPLUG-DocOwl, demonstrating the effectiveness of our model further.

\begin{table}[!h]
        \centering
	\begin{tabular}{llcccc}
        \toprule[.9pt]
        \multirow{3}{*}{\textbf{Type}} &\multirow{3}{*}{\textbf{Method}} & \multirow{3}{*}{\textbf{Size}}   &\multicolumn{3}{c}{RefCOCO} \\
         \cmidrule(rl){4-6}
            & & &\textbf{val} & \textbf{testA} & \textbf{testB}  \\
            \midrule
	\multirow{4}{*}{Traditional}	&OFA-L~\cite{wang2022ofa} & - &80.0&  83.7 & 76.4    \\
            &TransVG~\cite{deng2021transvg} & - & 81.0& 82.7 & 78.4  \\
            &VILLA~\cite{gan2020large} & - &  82.4  & 87.5 & 74.8 \\
            &UniTAB~\cite{yang2022unitab}  & -  & 86.3 &88.8 & 80.6 \\

             \midrule
             
        \multirow{5}{*}{LLM-based}    &VisionLLM-H~\cite{wang2023visionllm}   &  - & - & 86.7 & -  \\
            &Shikra-7B~\cite{chen2023shikra}   &  7B & 87.0 & 90.6 & 80.2  \\
            &Shikra-13B~\cite{chen2023shikra}   &  13B & 87.8 & 91.1 & 81.7  \\
            &Qwen-VL-chat~\cite{qwen}   &  7B & 88.6 & 92.3 & 84.5  \\
            &Next-chat~\cite{zhang2023next}   &  7B & 85.5 & 90.0 & 77.9  \\

            \midrule
            &Vary-toy & 1.8B &  88.1 & 90.6 & 85.7  \\
        \bottomrule[.9pt]
	\end{tabular}
        \setlength{\abovecaptionskip}{0.2cm}
        \vspace{1mm}
       \caption{Comparison with popular methods on RefCOCO. Benefiting from the new vision vocabulary, Vary-toy can achieve 88.1\% accuracy on RefCOCO val, which is on par with the 7B Qwen-VL-chat.}
        \label{tab:3}
        \vspace{-2mm}
\end{table}

\subsection{Natural Object Perception Ability}

The vision vocabulary network generated by Vary-tiny+ should enjoy two main advanced perception abilities: one for dense text and the other for natural objects. In this part, We test the latter ability of Vary-toy after accessing the improved vision vocabulary. It is worth noting that a center crop operation processes the input image of the CLIP branch. Therefore, it can be ruled out that the model uses CLIP for object localization. 

As shown in Table~\ref{tab:3}, Vary-toy can get 88.1\% accuracy@0.5 on the RefCOCO validation set, which is also on par with Qwen-VL-chat (7B) and even better than the Shikra-13B. The results show that under the knowledgeable vision vocabulary, Vary-toy gathers great natural object perception ability, proving the effectiveness of using the  Vary-tiny+ architecture to build a vision vocabulary, allowing us to further reflect on the necessity of CLIP if we add a large amount of weakly labeled image caption data, \textit{e.g.,} Laion-400M~\cite{schuhmann2021laion}, during the new vocabulary generating process.

\begin{table}[!h]
        \centering
	\begin{tabular}{lccccccc}
        \toprule[.9pt]
        \multirow{3}{*}{\textbf{Method}} & \multicolumn{7}{c}{MM-Vet}\\
        \cmidrule(rl){2-8}
             & \textbf{Rec} & \textbf{OCR} & \textbf{Know} & \textbf{Gen} & \textbf{Spat} & \textbf{Math} & \textbf{Total} \\
            \midrule
		BLIP-2~\cite{BLIP2}   &  27.5 & 11.1 & 11.8 & 7.0 & 16.2 & 5.8 & 22.4 \\
            LLaVA-7B~\cite{llava} & 28.0 & 17.1 & 16.3 & 18.9 & 21.2 & 11.5 & 23.8\\
            MiniGPT-4~\cite{minigpt4}  & 29.9 & 16.1 & 20.4 & 22.1 & 22.2 & 3.8 & 24.4 \\
            Otter~\cite{li2023otter} & 27.3 & 17.8 & 14.2 & 13.8 & 24.4 & 3.8 & 24.7 \\
            OpenFlamingo~\cite{Flamingo} & 28.7 & 16.7 & 16.4 & 13.1 & 21.0 & 7.7 & 24.8\\
            LLaVA1.5-7B~\cite{liu2023improvedllava} & - & - & - & -& - & - & 30.5 \\
            \midrule
            Vary-toy (1.8B) &  33.4 & 20.3 & 19.9 & 17.5 & 24 & 10.8    & 29.0 \\
        \bottomrule[.9pt]
	\end{tabular}
        \setlength{\abovecaptionskip}{0.2cm}
        \vspace{1mm}
       \caption{Comparison with popular LVLMs on MMVet. With only a 1.8B language model, Vary-toy can get a promising 29.0\% accuracy. The abbreviations represent Rec: Recognition; Know: Knowledge; Gen: Language generation; Spat: Spatial awareness.}
        \label{tab:4}
\end{table}

\subsection{General Ability}
We report the accuracy of Vary-toy on MMVet~\cite{yu2023mm} to test its general performance.  As summarized in table~\ref{tab:4},  Vary-toy gains 29.0\% of the total accuracy, which is higher than the classical LLaVA-7B~\cite{llava} (23.8\%), BLIP-2~\cite{BLIP2}, and MiniGPT-4~\cite{minigpt4}, demonstrating Vary-toy enjoys excellent general knowledge. For a 1.8B model, the generation ability is relatively poor and needs to be strengthened.

\begin{figure}[h]
\centering
\includegraphics[width=0.96\textwidth]{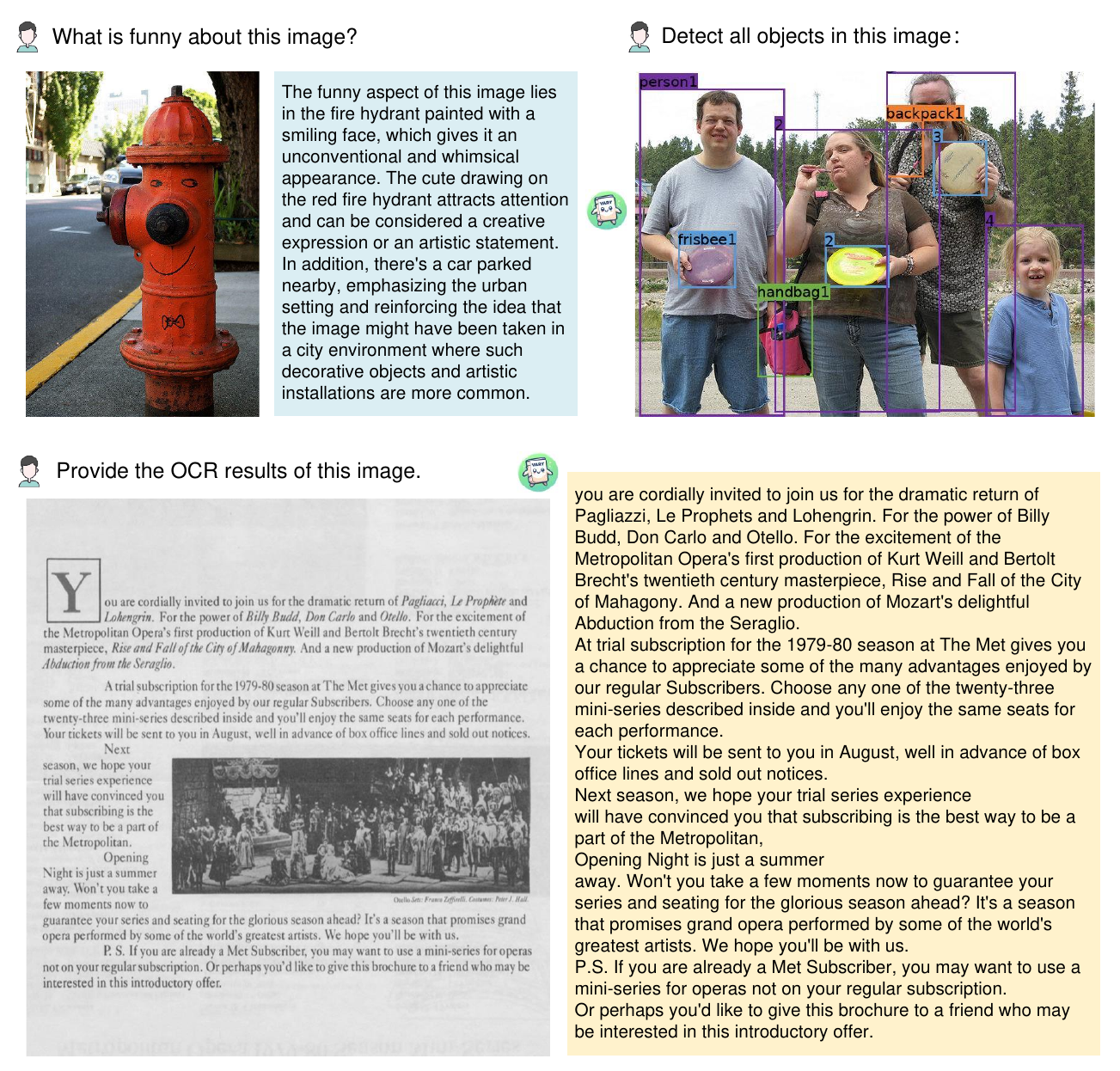}
\caption{Visualization of high-quality results of our model in four common fields. We can see that Vary-toy has satisfactory general ability and enjoys strong text and object perception abilities.}
\label{fig4}
\end{figure}
\vspace{-3mm}

\subsection{Visualization}
Figure~\ref{fig4} shows high-quality results of Vary-toy on four different downstream fields.  We can see that the model enjoys good vision concept understanding and localization capacities, indicating that a reinforced vision vocabulary with a small language model can also perform well in multimodal tasks.

\section{Conclusion}

In this report, we propose a small LVLM — Vary-toy, which can be deployed on a GTX1080ti GPU and enjoys fine performance in many downstream tasks. What's more, we generate a new and more comprehensive vision vocabulary for the presented model, which is the key to the success of Vary-toy. We hope the promising and user-friendly Vary-toy can become a new baseline in such fields as well as draw more attention to LVLM, especially for researchers who previously lacked computing resources. We also encourage researchers to use our reinforced vision vocabulary for more downstream tasks. Finally,  we firmly confirm that the Vary-toy will evolve beyond just a toy.

{
\small
\bibliographystyle{splncs04}
\bibliography{egbib}
}


\end{document}